\documentclass{article}
\usepackage{amsrefs}
\newenvironment{rezabib}
  {\bibdiv\biblist\setupbib}
  {\endbiblist\endbibdiv}

\def\setupbib{\catcode`@=\active}
\begingroup\lccode`~=`@
  \lowercase{\endgroup\def~}#1#{\gatherkey{#1}}
\def\gatherkey#1#2{\gatherkeyaux{#1}#2\gatherkeyaux}
\def\gatherkeyaux#1#2,#3\gatherkeyaux{\bib{#2}{#1}{#3}}

\usepackage{cvpr}
\usepackage{times}
\usepackage{epsfig}
\usepackage{graphicx}
\usepackage{amsmath}
\usepackage{amssymb}
\usepackage{amsmath}
\usepackage{graphicx}
\usepackage{tgtermes}

\usepackage[breaklinks=true,bookmarks=false]{hyperref}

\cvprfinalcopy 

\def\cvprPaperID{****} 

\begin{document}

\title{Unsupervised 3D Reconstruction from a Single Image via Adversarial Learning}

\author{Lingjing Wang\\
NYU Multimedia and Visual Computing Lab\\
Courant Institute of Mathematical Science\\
NYU Tandon School of Engineering, USA\\
{\tt\small lingjing.wang@courant.nyu.edu}
\and
Yi Fang \thanks{Corresponding author. Email: yfang@nyu.edu}\\
NYU Multimedia and Visual Computing Lab\\
Dept. of ECE, NYU Abu Dhabi, UAE\\
Dept. of ECE, NYU Tandon School of Engineering, USA\\
{\tt\small yfang@nyu.edu}
}


\maketitle

\begin{abstract}

Recent advancements in deep learning opened new opportunities for learning a high-quality 3D model from a single 2D image given sufficient training on large-scale data sets. However, the significant imbalance between available amount of images and 3D models, and the limited availability of labeled 2D image data (i.e. manually annotated pairs between images and their corresponding 3D models), severely impacts the training of most supervised deep learning methods in practice. In this paper, driven by a novel design of adversarial networks, we have developed an unsupervised learning paradigm to reconstruct 3D models from a single 2D image, which is free of manually annotated pairwise input image and its associated 3D model. Particularly, the paradigm begins with training an adaption network via autoencoder with adversarial loss, which embeds unpaired 2D synthesized image domain with real world image domain to a shared latent vector space. Then, we jointly train a 3D deconvolutional network to transform the latent vector space to the 3D object space together with the embedding process. Our experiments verify our network's robust and superior performance in handling 3D volumetric object generation from a single 2D image.

\end{abstract}

\section{Introduction}

3D object reconstruction from a single 2D image is a stimulating topic, being widely discussed in recent times in the 3D visual computing community. It has a wide range of applications related to  architecture,  visualization, e-commerce, 3D printing, etc. The severe loss of information from a 2D image to a 3D object makes the 3D reconstruction task quite challenging, but on the other hand, successful reconstruction of 3D object from a single 2D image becomes quite meaningful. 

Many efforts have been made to address this problem. Traditional methods\cites{haming2010structure, fuentes2015visual,rothganger2007segmenting,wu2013towards,wu2011visualsfm} achieved impressive results in the 3D reconstruction task. However, most of these methods require strong assumptions such as a dense number of views, high calibrated cameras, successful segmentation of objects from backgrounds, etc. Moreover, these methods are not applicable in the scenario where only a single 2D image is used as input. Nowadays, rapid developments in the field of deep learning and the availability of a large number of 3D CAD models\cites{chabot2016accurate,rock2015completing,bongsoo2015enriching} provide us the opportunity to approach this question via deep learning methods. 
However, the shortage is that deep learning methods usually require a large scale of data for training and we need to assume a strong similarity between training and testing data set. Let us briefly review related works below. 

\subsection{Related Works}
\subsubsection{3D Object Reconstruction Methods}
Existing works on 3D object reconstruction from 2D image(s) can be broadly categorized as two of the following: traditional methods without learning; deep learning based methods.

\noindent\textbf{3D reconstruction without learning.} 
The majority of traditional reconstruction methods based on SFM or SLAM\cites{haming2010structure, fuentes2015visual} are subject to a dense number of views, and most of them rely on the hypothesis that features can be matched across views. 2D to 3D reconstruction models such as multi-view stereo \cites{okutomi1993multiple,goesele2007multi}, space carving \cite{furukawa2006carved}, multiple moving object and large scale structure from motion \cites{rothganger2007segmenting,wu2013towards,wu2011visualsfm}, have all demonstrated good performance in solving the 2D to 3D reconstruction problem. However these methods require high calibrated cameras and segmentation of objects from their background, which are less applicable in practice. 

\noindent\textbf{Deep Neural Networks in 3D visual computing.} Nowadays, by generating 3D volumetric data\cite{wu20153d}, prominent deep learning models such as the deep 2D convolutional neural networks can be naturally extended to learn 3D objects. Deep learning models have proven to have strong capabilities in learning latent representative vector space of 3D objects \cite{wu20153d}. Multi-View CNN, Conv-DAE, Voxnet, Gift, T-L embedding, 3DGAN and so on, have uncovered great potential for solving retrieval, classification, 3D reconstruction problem, etc. on\cite{su2015multi,sharma2016vconv,maturana2015voxnet,bai2016gift,girdhar2016learning,3dgan}.

In contrast to the vast amount of research and accomplishments in the field of 3D object classification and retrieval, there are fewer research and far less accomplished results on 3D object reconstruction. Recently, researchers began to utilize 3D deconvolutional neural network to generate 3D volumetric objects from 2D images, for instance, 3D-GAN\cite{3dgan} and T-L embedding \cite{girdhar2016learning} strive to learn a latent vector space representation of 2D images, and then transform it to generate 3D volumetric objects. In addition, by leveraging LSTM based models, 3D-R2N2\cite{choy20163d} proposed a way to accumulate information from multiple-images for a better 3D object reconstruction. More recent research used 3D point clouds\cite{fan2016point} instead of volumetric data for a better 3D object reconstruction and achieved impressive performance.  
\vspace{-0.4cm}
\subsubsection{Adversarial Networks}
For 2D image generation, such as indoor scenes  \cite{yu2015lsun} and human faces  \cite{liu2015deep}, GAN based models, such as DCGAN\cite{radford2015unsupervised}, show impressive performance. Arjovsky et al proposed Wasserstein GAN (WGAN)   \cite{arjovsky2017wasserstein} with a more stable convergence. However, training GANs is challenging since the balance between generator's loss and discriminator's loss is hard to reconcile. BEGAN \cite{berthelot2017began} gives a solution to guarantee the equilibrium status, which decreased the difficulty in hyper-parameters setting for training. More recently, Cycle-GAN\cite{CycleGAN2017} is proposed to learn image transition via unpaired data and Fader Network \cite{lample2017fader} is proposed to slide attributes from image via autoencoder. However, there are very few researches that applied GANs in learning 3D objects. 3D-GAN and 3D-VAE-GAN proposed by Wu et al \cite{3dgan} is the first to adopt adversarial loss in 3D object reconstruction. 
\subsection{Our Solution}
Despite great success achieved in this field, there is a main challenge which is less discussed in previous works for 3D object reconstruction from a single image. On one hand, for most supervised methods, such as T-L embedding\cite{girdhar2016learning}, 3D-R2N2\cite{choy20163d}, 3DGAN\cite{3dgan} and 3D Point Clouds Method\cite{fan2016point}, robust performance requires a large number of labeled data for training. The limited amount of available labeled real world image data with their corresponding 3D objects dramatically reduces the efficiency of applying these models in practice. On the other hand, if we train the model using a large amount of paired synthesized images with their corresponding 3D objects, the trained model, in theory, can only predict 3D objects on real world images which are similar to the training synthesized data. 

\begin{figure}[h]
\centering
\includegraphics[width=9.5cm, height=6.5cm]{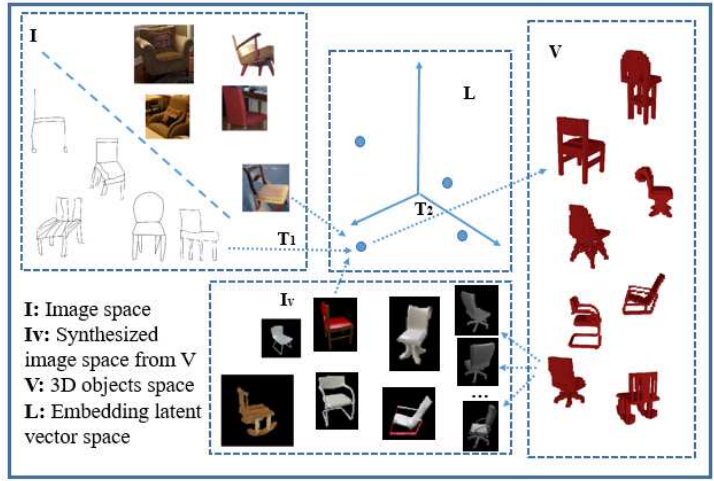}
\caption{Unsupervised space embedding for $I$ and $I_V$ to $L$ and transformation from space $L$ to $V$.}
\label{fig1}
\end{figure}

We propose a new question: what if we do not have any labeled images with corresponding 3D objects for training and the target testing images have totally different style from the possible training images (synthesized), such as for the human sketches? 
To clarify our question, we redefine the unsupervised 3D reconstruction problem. As shown in Figure \ref{fig1}, we assume that 2D images space $I=\{x_1,x_2,...,x_n\}$, where $x_i$ is a real world image, and 3D objects data space $V=\{v_1,...,v_m\}$, where $v_j$ is a 3D object. We call it "unsupervised 3D reconstruction" if we do not require any one-to-one paired connections between elements $x_i$ in space $I$ and elements $v_i$ in set $V$ for training. Since the images from $I$ are fully accessible, we allow $I$ to be used during the training process. Our target is to learn a transformation $T: I\rightarrow V$ to reconstruct the 3D objects for images in set $I$. We allow to render/synthesize images from 3D objects space $V$ and we call the synthesized image space $I_V$. Elements in $V$ and $I_V$ are fully paired. We assume the real world images from space $I$ are very different in style from the synthesized images from space $I_V$, without this assumption we can just apply supervised learning methods. 

We propose a novel adversarial autoencoder to learn a transformation $T_1: (I \cup I_V) \rightarrow L$, where $L\subset \mathbb{R}^n$, so that it can embed $I$ and $I_V$ onto a shared latent vector space $L$ and transformation $T_2 : L \rightarrow V$ to transform the latent vector space L to the 3D object space $V$ as shown in Figure \ref{fig1}. Transformation $T$, where $T=T_2T_1$ is the desired learning target. More specifically, transformation $T_1$ balances 2D reconstruction loss and adversarial loss, which means we need an ideal latent vector space $L=T_1(I)\cup T_1(I_V)$, whose vectors can not only represent the vector from input domain $I$ and $I_V$, but also non-distinguishable for vectors from space $I$ and vectors from space $I_V$. Transformation $T_2$ balances 3D reconstruction loss between 3D object and its rendered image and the adversarial loss to ensure element in $T_2(L)$ has similar distribution of 3D object in $V$. We continue use these notations throughout this paper. 

In sum, we list our main contributions below:

\begin{itemize}
\item We define a novel unsupervised 3D reconstruction from a single 2D image task, which is more appropriate for real world applications.

\item We propose a novel unsupervised paradigm to reconstruct 3D volumetric object on a single 2D image with embedding process. Our network can predict real world images, which are totally different in style from synthesized images, in an unsupervised way.

\item As an intermediate result, our model gives a new way to adapt two domains into one latent vector space, not requiring one-to-one pairwise data for supervision. 

\item We explored 3D reconstruction from a single human sketch image in an unsupervised way. Our model achieved superior performance in handling this task. 

\end{itemize}



\begin{figure*}[h]
\centering
\includegraphics[width=9.5cm, height=5.5cm]{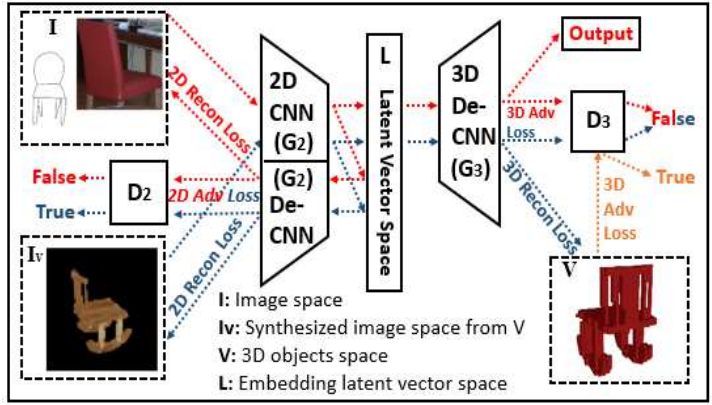}
\caption{Framework of the proposed method. The inputs of this system are image from space $I$ (unpaired with elements of $I_v$ or $V$), synthesized image from space $I_v$ with its corresponding 3D object from space $V$. Our target (output) is to reconstruct 3D object on image from space $I$. $D_2$ and $D_3$ are two discriminators for 2D image and 3D volumetric object. The generators are: $G_2$ and $G_3$, which reconstruct 2D images and 3D objects from the shared latent vector space $L$. $D_2$ treats the output $G_2(v)$ for $v\in I_V$ as ``True" and $G_2(w)$ for $w \in I$ as ``False". $D_3$ treats the $G_3(l) \text{ for } l \in L$ as ``False" and $G_3(v)$ for $v\in V$ as ``True". Generators balance reconstruction loss and adversarial loss during the training process. 2D loss and 3D loss are summed up for generator and discriminator.}
\label{fig2}
\end{figure*}

\section{Approach}
Our objective is to build an unsupervised generative neural network that accurately predicts 3D volumetric objects from a single real world 2D image. 
In general, as shown in Figure \ref{fig2}, our network includes two parts: Adapting 2D rendered images domain($I_V$) and real world images domain($I$) into one shared latent vector space($L$) via autoencoder($G_2$) with 2D adversarial loss and 2D reconstruction loss in section 2.1. Transformation($G_3$) from shared latent vector space($L$) to 3D objects space($V$) with 3D adversarial loss and 3D reconstruction loss in section 2.2. 
We explain the detailed training process in section 2.3. 

\subsection{Embedding real world image domain and synthesized image domain via 2D autoencoder}
In this section, we propose a novel approach based on 2D autoencoder to embed real world image domain and synthesized image domain. As shown in Figure \ref{fig2}, the autoencoder includes two parts,  $T_1\text{(2D CNN)}: I\cup I_v \rightarrow L$ and $T_1'\text{(De-CNN)}:L\rightarrow I\cup I_V$. Let 2D generator $G_2=T_1'T_1$. We utilize the following 2D reconstruction loss for $G_2$: 
\begin{equation}
L_{rec_2}(v|\theta_{G_2})=||T_1'T_1(v)-v||_1, \forall \text{ } v \in I \cup I_V. 
\end{equation}
where $\theta_{G_2}$ is the unknown weights in $G_2$. Let $D_2$ be a 2D discriminator to distinguish the distribution of $T_1'T_1(v)$, for $v\in I$ and  $T_1'T_1(w)$, for $w\in I_V$.  
BEGAN\cite{berthelot2017began} with an autoencoder based discriminator has shown more effective and robust performance than the traditional GANs. Training BEGAN is easier and we do not require the normal distribution assumption and KL divergence loss for adversarial network. Inspired by that, we define $L_2(v)=||v-D_2(v)||_{1} \text{ }\forall v \in T_1'T_1(I) \cup T_1'T_1(I_V)$, where $D_2$ is an autoencoder. We have the following objective function considering the adversarial loss according to BEGAN\cite{berthelot2017began}, $\forall \text{ } w \in I, \text{ } w_V \in I_V$,
\begin{equation}
\begin{split}
L_{D_2}(w,w_V|\theta_{D_2})&= L_2(T_1'T_1(w_V))-k_tL_2(T_1'T_1(w)),\\    
L_{G_2}'(w|\theta_{G_2})&=L_2(T_1'T_1(w))),\\ 
k_{t+1}(w,w_V):&=k_t(w,w_V)+\lambda_2(\gamma_2(L_2(T_1'T_1(w_V))\\
&-L_2(T_1'T_1(w)))
\end{split}
\end{equation}
where $\theta_{D_2}$ is the unknown parameters to be learned for $D_2$. Here, $\lambda_2$ is set to 0.01 and $\gamma_2$ is set to 1.15. 

By adding the reconstruction loss together, we can obtain the following loss for generator $G_{2}$: $\forall \text{ } w \in I, \text{ } w_V \in I_V$,
\begin{equation}
\begin{split}
L_{G_2}(w,w_V|\theta_{G_2})=&0.5\cdot(1-\phi_2)(L_{rec_2}(w)\\
&+L_{rec_2}(w_V))+\phi_2 L'_{G_2}(w) \\
\end{split}
\end{equation}

where $\phi_2$ is a hyper-parameter to balance the 2D reconstruction loss with 2D adversarial loss for $G_2$. The embedding performance is very sensitive for different $\phi_2$ and we will discuss it in section 3.3.3.





\subsection{Transformation from the shared latent vector space to 3D objects space}

The second part of our model is to learn a transformation $T_2 : L \rightarrow V$. $T_2$ is a typical 3D deconvolutional network, which can be regarded as a 3D generator $G_3$ as shown in Figure \ref{fig2}. Since we assume that we do not have corresponding 3D objects for the real world images, we can only define the 3D reconstruction loss on domain $I_V$,  $\forall w\in I_V, v_w\in V$ 
\begin{equation}
L_{rec_3}(w,v_w|\theta_{G_3})=||T_2(T_1(w))-v_w||_1,
\end{equation}
where $\theta_{G_3}$ denotes the unknown parameter of $G_3$.  

Similar to section 3.1, suppose $D_3$ be a discriminator to (a 3D-3D autoencoder) distinguish the distribution of $T_2(T_1(v))$ and $L_3(v)=||v-D_3(v)||_{1}$, we adopt the following loss function, $\forall w \in I, w_V \in I_V, v\in V$,

\begin{equation}
\begin{split}
L_{D_3}(v,w,w_V|\theta_{G_3})=&L_3(v)-0.5\cdot s_t\cdot(L_3(T_2T_1(w))\\
&+L_3(T_2T_1(w_V)),\\ 
L_{G_3}'(w,w_V| \theta_{G_2} , \theta_{G_3})=&0.5\cdot (L_3(T_2T_1(w))\\
&+L_3(T_2T_1(w_V)))  \\
s_{t+1}(w,w_V):=&s_t(w,w_V)+\lambda_3\cdot (\gamma_3\cdot (L_3(T_2T_1(w))\\
&-0.5\cdot(L_3(T_2T_1(w))\\
&+L_3(T_2T_1(w_V))),\\    
\end{split}
\end{equation}
where $\theta_{G_3}$ is the unknown parameter of $D_3$. $\lambda_3$ and $\gamma_3$ are set to 0.01 and 1.15, respectively.  

By adding the reconstruction loss, we can define the generation loss for $G_{3}$, $\forall w \in I, w_V \in I_V, v\in V$, 
\begin{equation}
\begin{split}
L_{G_3}(w,w_V,v| \theta_{G_2},\theta_{G_3}) =&(1-\phi_3)L_{rec_3}(w_V,v)\\
&+\phi_3 L'_{G_3}(w,w_V),
\end{split}
\end{equation}
where $\phi_3$ is a hyper-parameter to balance the 2D reconstruction loss with 2D adversarial loss for $G_3$. 

As a result, we have our overall loss functions:
\begin{equation}
\left\{\begin{array}{lr}   
L_{D}= L_{D_2}+L_{D_3},   \quad \textrm{for} \quad  \theta_{D2} \text{ and }  \theta_{D_3}\\ 
L_{G}=L_{G_2}+L_{G_3}, \quad \textrm{for} \quad  \theta_{G2} \text{ and }  \theta_{G_3}\\ 
\end{array}\right.
\end{equation}

\subsection{Training Process}

The network described above is trained using batches of data: $\{(w_i,{w_V}_i,v_i) | w_i \in I, {w_V}_i \in I_V, v_i \in V \}_{i=1,2,...,t}$. t is the batch size and we set it equal to 32. The images $I_V$ are rendered from 3D models $V$. More specifically, all the models are rendered into 24 views from 0 to 360 degrees. Please refer 3D-R2N2\cite{choy20163d} for detailed information. We directly use their prepared rendered data. Our codes are written using Tensorflow.

We train our model in two steps. We pre-train the 2D autoencoder to adapt real world image domain with synthesized image domain. The input are unpaired single 2D image and synthesized single 2D image with dimension 64x64. We initialize the network at random. We set the learning rate starting from 0.005 with 0.995 exponential decay for the Adam optimizer for $G_2$, and learning rate starting from 0.001 with 0.995 exponential decay for the Adam optimizer for $D_2$. We use leaky-relu\cite{xu2015empirical} activation function. We implement batch normalization\cite{ioffe2015batch} on each layers except for output layer. The 2D reconstruction loss is L-1 norm between input 2D image and reconstructed 2D image. We train this for about 100k steps. Then, we fix the weights in $G_2$ to train $G_3$ and $D_3$. The input for $G_3$ is the latent vector from $L$. We initialize the network at random. We use the same settings for $G_3$ and $D_3$ exactly as for $G_2$ and $D_2$. The 3D reconstruction loss is L-1 norm between 3D ground truth object and reconstructed 3D object from synthesized images. We train this for about 100k steps and then we start to train the whole system with embedding process using loss function (7) in section 2.3 until convergence.

\section{Experiments}
\subsection{Data sets and Evaluation Settings}
We use the following three benchmark data sets for testing the performance of the proposed method on 3D object reconstruction: \textbf{ShapeNet}\cite{chang2015shapenet}, \textbf{PASCAL 3D}\cite{xiang2014beyond}, and \textbf{SHREC 13}\cite{li2013shrec}.  

\noindent\textbf{ShapeNet: } \textbf{ShapeNet} is a large scale organized collection of richly-annotated 3D CAD models \cite{chang2015shapenet}. We directly used the subset data with rendered images from 3D-R2N2\cite{choy20163d}. Their data set provides us 3D volumetric objects in 13 classes, but we only use those classes which appears in the following real world image data sets. 

\noindent\textbf{PASCAL 3D: } \textbf{PASCAL 3D} data set augmented 3D CAD models on PASCAL 2012 detection images \cite{xiang2014beyond}. We chose a subset of PASCAL 3D that contains 3000 models covering 4 categories. The 2D images in this data set are real world images with complex background and many of the images are cropped from a larger image, containing only a small part of the actual object. 

\noindent\textbf{SHREC 13: } The \textbf{SHREC 13} data set includes 20000 human-drawn sketches, categorized into 250 classes, each with 80 sketches. There are 10 categories in \textbf{SHREC 13} data set which are included in our 13 classes prepared \textbf{ShapeNet} data. The sketch images do not have corresponding 3D objects. We use all of the 80 sketches. In this paper, we only use the "chair category" for demonstration.  

\noindent\textbf{Evaluation Metric.} Our volumetric data has the shape 32x32x32.
We use Intersection-over-Union (IoU) between 3D ground truth and our 3D voxel reconstruction to evaluate our network. More specifically,
\begin{equation}
IoU =\frac{\sum_{i,j,k}[I(p_{(i,j,k)}>t)I(y_{(i,j,k)})]}{\sum_{i,j,k}[I(I(p_{(i,j,k)}>t) +I(y_{(i,j,k)})]}
\end{equation}
where $t$ is a voxelization threshold that we set to equal to 0.3 (This value is insensitive to the calculated IoU in our case), $I(\cdot)$ is an indicator function. Better 3D reconstruction results are usually associated with higher IoU values.

\subsection{Learning the embedding latent vector space for images and synthesized images}
The embedding process is the key and the most difficult part in our model. Unsuccessful embedding result will cause a totally wrong reconstructed 3D object, which can be completely different from the target. However, due to the big difference between synthesized image and real world/sketch image, and the fact that we do not have paired information for training, the embedding process becomes very challenging. In this section, we train the 2D autoencoder($G_2$) with 2D reconstruction and adversarial loss to embed the image space with synthesized image space. The inputs are unpaired synthesized image and real world/human sketch image and the output is vectors in the shared latent vector space. For our analysis, we check the reconstructed 2D image from the autoencoder, even though they do not have direct impact on our final 3D reconstruction result.

\subsubsection{Experimental Setting} 
In this section, we use images from \textbf{PASCAL} and \textbf{SHREC} and 3D objects from \textbf{ShapeNet} for training, and the same images for testing. The 3D objects data with synthesized images are split into 7:3 ratio for training and testing for section 3.2.3 only. We use the chair category for demonstration. The model is separately trained for \textbf{PASCAL} images and \textbf{SHREC} images. 

For convenience, we use the notation $C(k,s)$ and $DC(k,s)$ to denote the convolutional and deconvolutional process in our network, respectively. $k$ is the number of kernels and s is the kernel shape ($s\times s$ for 2D convolution and $s\times s\times s$ for 3D convolution). The details of our network structure for these experiments are demonstrated below. 2D-2D autoencoder: $C1(32,4)-C2(64,4)-C3(128,4)-FC1(512)-FC2(200)-FC3(512)-DC1(128,4)-DC2(64,4)-DC3(32,4)$. Since our discriminator "$D_2$" is also an autoencoder, we use exactly the same structure with different weights. For details of the training process please refer to      ``Training Process" in Section 2.3. All experiments are conducted on a server with one K80 GPU. The maximal iteration numbers are set to approximately 150k. 
\begin{figure}[h]
\centering
\includegraphics[width=8.5cm, height=5.5cm]{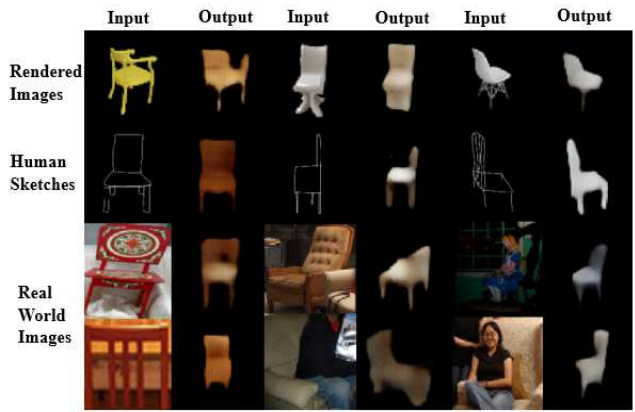}
\caption{The outputs of the 2D autoencoder when $\phi_2=0.7$.}
\label{fig2}
\end{figure}

\begin{figure}[h]
\centering
\includegraphics[width=7.5cm, height=8cm]{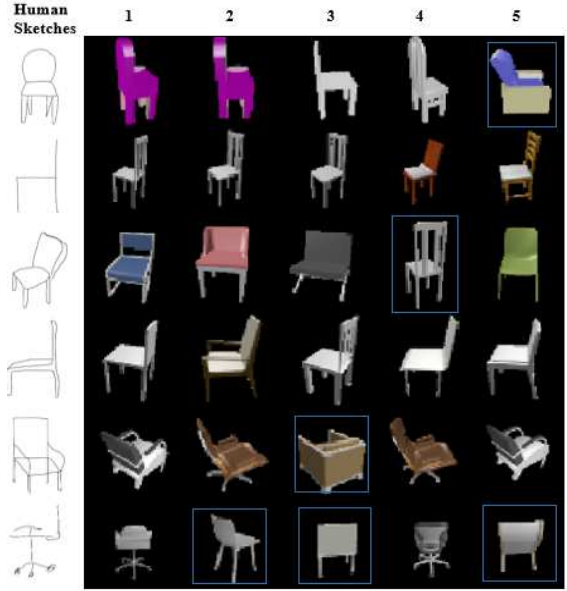}
\caption{Synthesized data retrieval from sketches}
\label{fig232}
\end{figure}

\subsubsection{Embedding performance: outputs from the 2D autoencoder}

In this section, we check the outputs of the 2D autoencoder for both synthesized images and real/sketch images to check the embedding performance. We randomly pick several observations presented in Figure \ref{fig2}. Since the embedding performance is very sensitive to parameter $\phi_2$, during the training process, we fix $\phi_2=0.7$. Firstly, from the input columns, we can clearly see the difference in style among rendered images, sketches and real world images, but the output shows the consistence in style among these three types of data. The second part we care about is information loss during the embedding process. Despite minor information losses, such as the detailed information of chairs' legs, the major part of the chair can be transformed precisely in the output after embedding, especially for rendered images and sketches. For real world images, most background information is filtered, but if the background has similar color as the object then the background information may be merged with the object during the transformation process. Besides, parts with very dark colors in the input image will be easily ignored, e.g. second example in row 4. But some missing parts can be reconstructed as a result of ``imagination" due to the adversarial loss such as the legs of second sofa chair in row 3.

\subsubsection{Embedding performance: retrieval results}

After training the autoencoder and learned the shared latent vector space, we check whether or not our latent vector space can still be a good representative descriptor after embedding. We use sketches and rendered images for demonstration in this section.  

For this part, we use around 1500 testing chairs' rendered image pool. These chairs were not used for training. We select sketches to check the first 5 closest rendered images from the pool by calculating the $L_2$ distance in their descriptor from the latent vector space. We pick out 6 different types of chairs and illustrate their retrieval performance in Figure \ref{fig232}. Column 1 shows the nearest neighbor to the sketches. In Figure \ref{fig232}, we see that for most chairs, their nearest neighbor are very close to the sketches. Failed cases are marked in blue boxes. In the failed cases, the chair's leg is the main problem. It is not surprising to see that the hidden layer of the CNN based autoencoder in the middle ignores subtle information of the input sketches.  

\subsubsection{Embedding performance: the functions of $\phi_2$}
\begin{figure}[h]
\centering
\includegraphics[width=9cm, height=5.5cm]{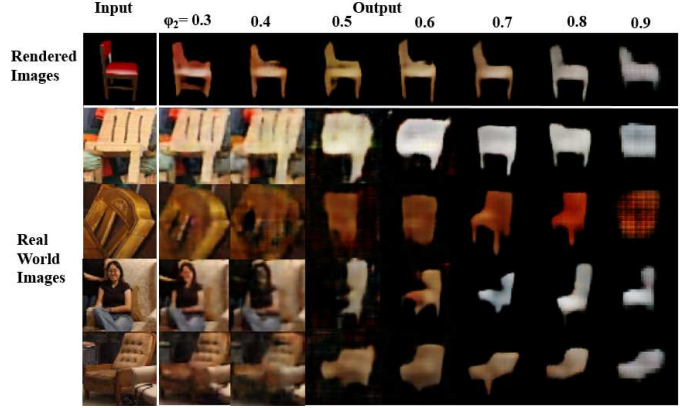}
\caption{Comparison of the outputs of 2D autoencoder after embedding process for different $\phi_2$}
\label{fig1132}
\end{figure}

From formula (3) in section 2.1, we know $\phi_2$ is a hyper-parameter to balance the 2D reconstruction loss and adversarial loss during the embedding process. In this part, we train our model for different $\phi_2$. This experiment gives us an opinion on the effects of $\phi_2$ and it demonstrates the robustness of the embedding process as well. In Figure \ref{fig1132}, we gradually see the difference among outputs for different $\phi_2$. When $\phi_2$ equals to 0.3, the output contains most background information and the difference between input and output are subtle. While $\phi_2$ increases, reconstruction loss becomes less important and adversarial loss starts to dominate the total loss. In this case, the difference in style between real images and synthesized images decreases and the difference between input and output increases. 
When $\phi_2 = 0.9$, shown in last column in Figure \ref{fig1132}, the output becomes very abstract and we can hardly distinguish the real output with synthesized images. 

\begin{figure*}[h]
\centering
\includegraphics[width=14cm, height=4.5cm]{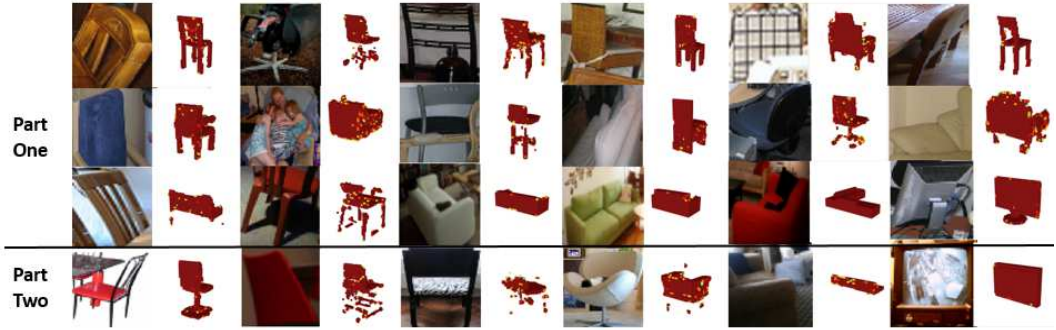}
\caption{Selected 3D reconstructed models for single real world image}
\label{fig11323}
\end{figure*}

\subsection{Reconstruction of 3D models}
In this section, we illustrate the 3D reconstruction network from a single image with embedding process. In 4.3.1, we test our model using Pascal 3D data. Based on the output (shared latent vector space $L$) from previous step, we train the 3d generator ($G_3$) and add a 3d discriminator ($D_3$) together for training. In 4.3.2, we compare our model with the model that excluded the embedding process. In 4.3.3, we test our model on human sketches. 

\subsubsection{Experimental setting.} 
Unlike classical supervised models which usually test the model performance using testing synthesized data, we use real world image data \textbf{PASCAL} and human sketches data \textbf{SHREC 13}, which are different in style from synthesized data for testing. Even thought most state-of-the-art supervised models did test their model on real world images, for example, 3D-R2N2\cite{choy20163d} tested their model on \textbf{PASCAL} data set, T-L embedding\cite{girdhar2016learning} and 3D-GAN\cite{3dgan} tested their model on \textbf{IKEA}\cite{lim2013sketch} data sets, but these models all require a fine-tuning process involving part of labeled \textbf{PASCAL} or \textbf{IKEA} images for training, which is totally different task. 3D Point Clouds Method\cite{fan2016point} tested their model on their own real world images without fine-tuning process. But their selected images include objects with complete shape and have similar style to synthesized data. Moreover, the background is masked for their testing. Therefore, due to these different task settings, our main target in this section is not to compare our method with the state-of-the-art supervised methods, but to demonstrate the unsupervised reconstruction performance of our model.   

In this section, we reuse the pre-training weights for layers from $C_1$ to $F_2$ of 2D autoencoder from the previous section. The discriminator $D_2$ remains the same as the previous section. In addition, we have a new 3D deconvolution from the latent vector space to 3D object space. A 3D discriminator $D_3$, which is a 3D-3D autoencoder, is added together. 

2D-3D Network: $C1(32,4)-C2(64,4)-C3(128,4)-FC1(512)-FC2(200)-FC3(512)-DC1^{3D}(128,4)-DC2^{3D}(64,4)-DC3^{3D}(32,4)$

Discriminator $D_3$: $C1^{3D}(32,4)-C2^{3D}(64,4)-C3^{3D}(128,4)-FC1(512)-FC2(200)-FC3(512)-DC1^{3D}(128,4)-DC2^{3D}(64,4)-DC3^{3D}(32,4)$

We set $\phi_2=0.7$ and $\phi_3=0.2$. We train it for approximately 100k steps with fixed weights of $G_2$ and then continue train the networks with embedding process until convergence on one K80 GPU. For 3.3.2 and 3.3.4, we show our reconstruction performance on real world and human sketches. For experiment in section 3.3.3, we use the same 2D-3D Network with reconstruction loss only, and train it from random weights for approximately 300k steps for single category and with other 9 classes together as well. The synthesized images with 3D objects data are split with ratio 7:3 for training and testing for this section. 

\subsubsection{3D reconstruction for real world images}
In this section, we test our model performance on the real world images. Even though the images have corresponding 3D models, we only use them for quantifying our result. 

\textbf{Qualitative Result: } In figure \ref{fig11323}, we select those models that are more representative than others. Chair is one of the most difficult categories. Compared to other categories, the complexity of the Chair's shape and subtle details are much higher. Therefore, it is more interesting to analyze the reconstruction of 3D chairs. In part one, we see that most of the reconstructed models captured the main information from the 2D image, such as the shape of the chairs' back. Some models even captured the correct information on the legs of the chairs, e.g. second chair in row 1 and row 3, and third chair in row 2. Unlike most supervised models that usually shows less complexity in their reconstructed 3D models after using part of the testing data for fine-tuning, our reconstructed model shows high complexity. In part two, we selected several typical failed cases. Failed cases in our result have reconstructed a perfect model that is unlike the chair in the input image, such as the first chair in part two. The reason for this failed case can be attributed to the fact that the color of the desk's leg is the same as the chair, and together looks like a stand. The second chair shows another type of failed case, which have missing components. This type of failure is less common since most of the missing components in the image should be recovered after adding adversarial loss. The third chair shows another major failure in our result because dark color in the image can be easily ignored during the reconstruction process.  
\begin{table}[h]
\begin{center}
\begin{tabular}{ cccccc} 
\hline
Method   &Car & Chair & Sofa & Tv Monitor\\
\hline
Kar et al\cite{kar2015category}& 0.472 & 0.234 & 0.149 & 0.492 \\
3D-R2N2-1\cite{choy20163d}   &0.579 & 0.203 & 0.251 & 0.438 \\ 
3D-R2N2-2\cite{choy20163d}   & 0.699 & 0.280 & 0.332 & 0.574 \\
Ours(unsupervised) &   0.634 & 0.241 & 0.45 & 0.247 \\ 
\hline
\end{tabular}
\end{center}
\caption{Per-Category IoU on Pascal 3D. Our results are computed using the entire Pascal 3D Data as testing data set. 3D-R2N2-1 stands for LSTM-1 version and 3D-R2N2-2 stands for Res3DGRu3 version. In comparison, 3D-R2N2 used partial Pascal data for fine-tuning. For details, please refer their papers \cite{choy20163d}.}
\label{tbl32}
\end{table}

\begin{table}[h]
\begin{center}
\begin{tabular}{ cccccc} 
\hline
Method  &IoU \\
\hline
3D R2N2(1-view)\cite{choy20163d} (synthesized test data)& 0.466 \\
3D Point Clouds Method \cite{fan2016point}(synthesized test data) & 0.544\\
2D-3D-1 without Adaption (synthesized test data)& 0.471  \\
2D-3D-2 without Adaption (synthesized test data)& 0.539  \\
2D-3D-2 without Adaption (Pascal 3D test data)& 0.153  \\
Ours (Pascal 3D test data) &  0.241 \\ 
\hline
\end{tabular}
\end{center}
\caption{IoU on different testing set for category chair. 2D-3D-1 is jointly trained with other 9 classes and 2D-3D-2 is trained separately for chairs only.}
\label{tbl321}
\end{table}

\begin{figure*}[h]
\centering
\includegraphics[width=14cm, height=5.5cm]{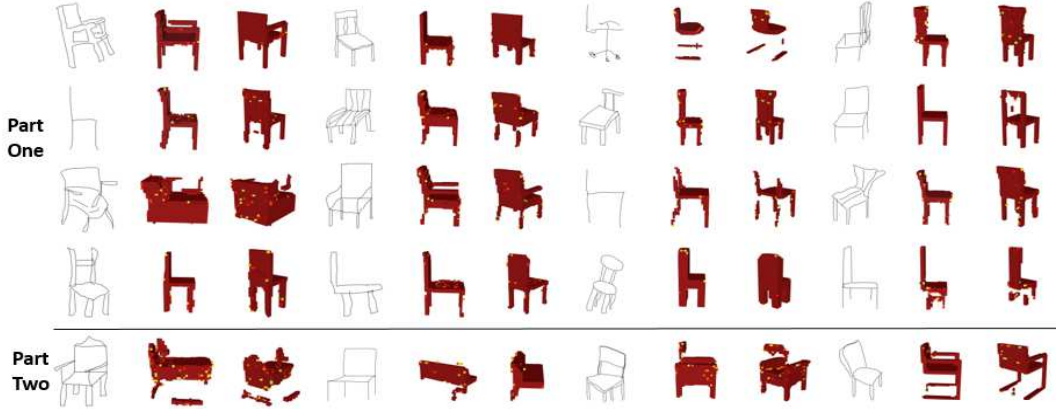}
\caption{Selected 3D reconstructed models from single human sketches}
\label{fig11323}
\end{figure*}

\textbf{Quantitative Result: }
In Table \ref{tbl32}, we list our results with previous state-of-the-art methods. Due to the differences in experimental setting (see 3.3.1), the results are provided for reference only. For category chair, we allow our reconstructed 3D model to be shifted and scaled to calculate IoU. For the monitor category, the testing data mainly include TV monitors, but our training data mainly include thin computer monitors, thus is why our IoU for monitor category is lower than Kar et al \cite{kar2015category} and 3D-R2N2 LSTM-1 \cite{choy20163d}. For category Car, Chair, we archived similar results as 3D-R2N2 and for category  Sofa, our model even achieved better result without supervised learning.

\subsubsection{Comparison of reconstruction results with/without embedding process}
In this section, 2D-3D supervised network is trained without embedding process using synthesized data. We apply the trained model to predict real world images directly. In Table \ref{tbl321}, we show that the simple 2D-3D supervised network achieves impressive results on synthesized testing data with 0.471 IoU and 0.539 IoU if separately trained. We use category chair for demonstration. However, the IoU of this model without adaption on \textbf{PASCAL} testing data is only 0.153 compared to our IoU 0.241. Moreover, the reconstructed 3D objects without embedding are noisy cloud points in many cases. In our opinion, this is not a typical over-fitting problem but the testing data is very different. 
\subsubsection{3D reconstruction for human sketches}
In this section, we try to use our model to predict 3D objects from a single human sketch. Due to the absence of labeled ground truth 3D models, there is no supervised methods that can be used for this task. To the best of our knowledge, no previous research explored this task. 

In Figure \ref{fig11323}, we show the 3D reconstruction result from human sketches. Since the sketch data for category chair only contains 80 images, we checked the 3D reconstructed model for each one. We select representative models with satisfactory performance in part one and failed examples in Part two. The failed ratio of total sample is less than 15 percent approximately. In part one, we see that the general shape of most sketches can be reconstructed correctly. For some models such as the first chair, more subtle information such as the characteristics of arms/legs can be seen in the 3D reconstructed model.  However, more tiny characteristics such as the special design on the back of the chair (the second example in row 2, and the third example in row 4) cannot be correctly reconstructed in most cases. In part two, we show several representative failed cases. Our model failed to reconstruct legs and arms for the first example. The reconstructed 3D models cannot capture the correct shape for the chair in the second example. The third and fourth chair examples seem to have learned the wrong image. Especially in the fourth example, it looks like a perfect reconstruction, however it is quite different from the sketch. The same sketch is shown in Figure \ref{fig232}, if we take a closer look, we can see its nearest neighbor has similar legs as in the reconstructed 3D model. Therefore, we make a reasonable hypothesis that the reason we have a totally different reconstructed model is due to the unsuccessful embedding.

\section{Conclusion}
Our paper addresses the challenging problem of predicting 3D objects from a single real world/human sketch image. Since most real world images such as human sketches do not have corresponding labels for training, and differ hugely with synthesized images in style, the development of unsupervised network for this task becomes crucial. Moreover, the same situation exists in many other fields of 3D visual computing, such as 3D pose estimation, there are fewer well-labeled real world data. Our proposed unsupervised network with embedding process via adversarial network is a possible solution that can be extended to explore other works such as pose estimation, retrieval, and classification under similar situations. Based on our model, we completed tasks such as 3D object prediction from a human sketch image, which to the best of our knowledge, no previous researches explored, due to the absence of corresponding 3D labels. We demonstrate superior and robust performance of our model in handling unsupervised 3D reconstruction task. 

\begin{rezabib}
@article{haming2010structure,
  title={The structure-from-motion reconstruction pipeline--a survey with focus on short image sequences},
  author={H{\"a}ming, Klaus and Peters, Gabriele},
  journal={Kybernetika},
  volume={46},
  number={5},
  pages={926--937},
  year={2010},
  publisher={Institute of Information Theory and Automation AS CR}
}
@article{fuentes2015visual,
  title={Visual simultaneous localization and mapping: a survey},
  author={Fuentes-Pacheco, Jorge and Ruiz-Ascencio, Jos{\'e} and Rend{\'o}n-Mancha, Juan Manuel},
  journal={Artificial Intelligence Review},
  volume={43},
  number={1},
  pages={55--81},
  year={2015},
  publisher={Springer}
}
@article{rothganger2007segmenting,
  title={Segmenting, modeling, and matching video clips containing multiple moving objects},
  author={Rothganger, Fred and Lazebnik, Svetlana and Schmid, Cordelia and Ponce, Jean},
  journal={IEEE transactions on pattern analysis and machine intelligence},
  volume={29},
  number={3},
  year={2007},
  publisher={IEEE}
}
@inproceedings{wu2013towards,
  title={Towards linear-time incremental structure from motion},
  author={Wu, Changchang},
  booktitle={3DTV-Conference, 2013 International Conference on},
  pages={127--134},
  year={2013},
  organization={IEEE}
}

@article{wu2011visualsfm,
  title={VisualSFM: A visual structure from motion system},
  author={Wu, Changchang and others},
  year={2011}
}
@inproceedings{chabot2016accurate,
  title={Accurate 3D car pose estimation},
  author={Chabot, Florian and Chaouch, Mohamed and Rabarisoa, Jaonary and Teuli{\`e}re, C{\'e}line and Chateau, Thierry},
  booktitle={Image Processing (ICIP), 2016 IEEE International Conference on},
  pages={3807--3811},
  year={2016},
  organization={IEEE}
}

@inproceedings{rock2015completing,
  title={Completing 3D object shape from one depth image},
  author={Rock, Jason and Gupta, Tanmay and Thorsen, Justin and Gwak, JunYoung and Shin, Daeyun and Hoiem, Derek},
  booktitle={Proceedings of the IEEE Conference on Computer Vision and Pattern Recognition},
  pages={2484--2493},
  year={2015}
}
@inproceedings{bongsoo2015enriching,
  title={Enriching object detection with 2d-3d registration and continuous viewpoint estimation},
  author={Bongsoo Choy, Christopher and Stark, Michael and Corbett-Davies, Sam and Savarese, Silvio},
  booktitle={Proceedings of the IEEE Conference on Computer Vision and Pattern Recognition},
  pages={2512--2520},
  year={2015}
}
@article{okutomi1993multiple,
  title={A multiple-baseline stereo},
  author={Okutomi, Masatoshi and Kanade, Takeo},
  journal={IEEE Transactions on pattern analysis and machine intelligence},
  volume={15},
  number={4},
  pages={353--363},
  year={1993},
  publisher={IEEE}
}

@inproceedings{goesele2007multi,
  title={Multi-view stereo for community photo collections},
  author={Goesele, Michael and Snavely, Noah and Curless, Brian and Hoppe, Hugues and Seitz, Steven M},
  booktitle={Computer Vision, 2007. ICCV 2007. IEEE 11th International Conference on},
  pages={1--8},
  year={2007},
  organization={IEEE}
}
@article{furukawa2006carved,
  title={Carved visual hulls for image-based modeling},
  author={Furukawa, Yasutaka and Ponce, Jean},
  journal={Computer Vision--ECCV 2006},
  pages={564--577},
  year={2006},
  publisher={Springer}
}
@inproceedings{wu20153d,
  title={3d shapenets: A deep representation for volumetric shapes},
  author={Wu, Zhirong and Song, Shuran and Khosla, Aditya and Yu, Fisher and Zhang, Linguang and Tang, Xiaoou and Xiao, Jianxiong},
  booktitle={Proceedings of the IEEE Conference on Computer Vision and Pattern Recognition},
  pages={1912--1920},
  year={2015}
}
@inproceedings{su2015multi,
  title={Multi-view convolutional neural networks for 3d shape recognition},
  author={Su, Hang and Maji, Subhransu and Kalogerakis, Evangelos and Learned-Miller, Erik},
  booktitle={Proceedings of the IEEE international conference on computer vision},
  pages={945--953},
  year={2015}
}
@inproceedings{sharma2016vconv,
  title={Vconv-dae: Deep volumetric shape learning without object labels},
  author={Sharma, Abhishek and Grau, Oliver and Fritz, Mario},
  booktitle={Computer Vision--ECCV 2016 Workshops},
  pages={236--250},
  year={2016},
  organization={Springer}
}
@inproceedings{maturana2015voxnet,
  title={Voxnet: A 3d convolutional neural network for real-time object recognition},
  author={Maturana, Daniel and Scherer, Sebastian},
  booktitle={Intelligent Robots and Systems (IROS), 2015 IEEE/RSJ International Conference on},
  pages={922--928},
  year={2015},
  organization={IEEE}
}
@inproceedings{bai2016gift,
  title={Gift: A real-time and scalable 3d shape search engine},
  author={Bai, Song and Bai, Xiang and Zhou, Zhichao and Zhang, Zhaoxiang and Jan Latecki, Longin},
  booktitle={Proceedings of the IEEE Conference on Computer Vision and Pattern Recognition},
  pages={5023--5032},
  year={2016}
}
@inproceedings{girdhar2016learning,
  title={Learning a predictable and generative vector representation for objects},
  author={Girdhar, Rohit and Fouhey, David F and Rodriguez, Mikel and Gupta, Abhinav},
  booktitle={European Conference on Computer Vision},
  pages={484--499},
  year={2016},
  organization={Springer}
}
@inproceedings{3dgan,
  title={Learning a probabilistic latent space of object shapes via 3d generative-adversarial modeling},
  author={Wu, Jiajun and Zhang, Chengkai and Xue, Tianfan and Freeman, William T and Tenenbaum, Joshua B}
  booktitle={Advances in Neural Information Processing Systems},
  pages={82--90},
  year={2016}
}
@inproceedings{choy20163d,
  title={3d-r2n2: A unified approach for single and multi-view 3d object reconstruction},
  author={Choy, Christopher B and Xu, Danfei and Gwak, JunYoung and Chen, Kevin and Savarese, Silvio},
  booktitle={European Conference on Computer Vision},
  pages={628--644},
  year={2016},
  organization={Springer}
}
@article{fan2016point,
  title={A point set generation network for 3d object reconstruction from a single image},
  author={Fan, Haoqiang and Su, Hao and Guibas, Leonidas},
  journal={arXiv preprint arXiv:1612.00603},
  year={2016}
}
@article{yu2015lsun,
  title={Lsun: Construction of a large-scale image dataset using deep learning with humans in the loop},
  author={Yu, Fisher and Seff, Ari and Zhang, Yinda and Song, Shuran and Funkhouser, Thomas and Xiao, Jianxiong},
  journal={arXiv preprint arXiv:1506.03365},
  year={2015}
}

@inproceedings{liu2015deep,
  title={Deep learning face attributes in the wild},
  author={Liu, Ziwei and Luo, Ping and Wang, Xiaogang and Tang, Xiaoou},
  booktitle={Proceedings of the IEEE International Conference on Computer Vision},
  pages={3730--3738},
  year={2015}
}

@article{radford2015unsupervised,
  title={Unsupervised representation learning with deep convolutional generative adversarial networks},
  author={Radford, Alec and Metz, Luke and Chintala, Soumith},
  journal={arXiv preprint arXiv:1511.06434},
  year={2015}
}
@article{arjovsky2017wasserstein,
  title={Wasserstein gan},
  author={Arjovsky, Martin and Chintala, Soumith and Bottou, L{\'e}on},
  journal={arXiv preprint arXiv:1701.07875},
  year={2017}
}
@article{berthelot2017began,
  title={BEGAN: Boundary Equilibrium Generative Adversarial Networks},
  author={Berthelot, David and Schumm, Tom and Metz, Luke},
  journal={arXiv preprint arXiv:1703.10717},
  year={2017}
}
@article{CycleGAN2017,
  title={Unpaired Image-to-Image Translation using Cycle-Consistent Adversarial Networks},
  author={Zhu, Jun-Yan and Park, Taesung and Isola, Phillip and Efros, Alexei A},
  journal={arXiv preprint arXiv:1703.10593},
  year={2017}
}

@article{lample2017fader,
  title={Fader Networks: Manipulating Images by Sliding Attributes},
  author={Lample, Guillaume and Zeghidour, Neil and Usunier, Nicolas and Bordes, Antoine and Denoyer, Ludovic and Ranzato, Marc'Aurelio},
  journal={arXiv preprint arXiv:1706.00409},
  year={2017}
}
@inproceedings{ioffe2015batch,
  title={Batch normalization: Accelerating deep network training by reducing internal covariate shift},
  author={Ioffe, Sergey and Szegedy, Christian},
  booktitle={International Conference on Machine Learning},
  pages={448--456},
  year={2015}
}

@article{xu2015empirical,
  title={Empirical evaluation of rectified activations in convolutional network},
  author={Xu, Bing and Wang, Naiyan and Chen, Tianqi and Li, Mu},
  journal={arXiv preprint arXiv:1505.00853},
  year={2015}
}

@article{chang2015shapenet,
  title={Shapenet: An information-rich 3d model repository},
  author={Chang, Angel X and Funkhouser, Thomas and Guibas, Leonidas and Hanrahan, Pat and Huang, Qixing and Li, Zimo and Savarese, Silvio and Savva, Manolis and Song, Shuran and Su, Hao and others},
  journal={arXiv preprint arXiv:1512.03012},
  year={2015}
}
@inproceedings{xiang2014beyond,
  title={Beyond pascal: A benchmark for 3d object detection in the wild},
  author={Xiang, Yu and Mottaghi, Roozbeh and Savarese, Silvio},
  booktitle={Applications of Computer Vision (WACV), 2014 IEEE Winter Conference on},
  pages={75--82},
  year={2014},
  organization={IEEE}
}
@book{li2013shrec,
  title={SHREC’13 track: large scale sketch-based 3D shape retrieval},
  author={Li, Bo and Lu, Yijuan and Godil, Afzal and Schreck, Tobias and Aono, Masaki and Johan, Henry and Saavedra, Jose M and Tashiro, Shoki},
  year={2013}
}
@inproceedings{lim2013sketch,
  title={Sketch tokens: A learned mid-level representation for contour and object detection},
  author={Lim, Joseph J and Zitnick, C Lawrence and Doll{\'a}r, Piotr},
  booktitle={Proceedings of the IEEE Conference on Computer Vision and Pattern Recognition},
  pages={3158--3165},
  year={2013}
}
@inproceedings{kar2015category,
  title={Category-specific object reconstruction from a single image},
  author={Kar, Abhishek and Tulsiani, Shubham and Carreira, Joao and Malik, Jitendra},
  booktitle={Proceedings of the IEEE Conference on Computer Vision and Pattern Recognition},
  pages={1966--1974},
  year={2015}
}

\end{rezabib}

\end{document}